%% file: main_arxiv.tex
\documentclass[runningheads]{llncs}
\usepackage{graphicx}
\usepackage{comment}
\usepackage{amsmath,amssymb} 
\usepackage{color}

\usepackage{times}
\usepackage{epsfig}
\usepackage{epstopdf}
\usepackage{capt-of}
\usepackage{booktabs}
\usepackage{array}
\usepackage{tabularx}
\usepackage{multirow}
\usepackage{booktabs}
\usepackage{multirow}
\usepackage[table,xcdraw]{xcolor}
\usepackage{siunitx}
\usepackage{gensymb}

\newcommand{\nelson}[1]{\textcolor{cyan}{}}
\newcommand{\chen}[1]{\textcolor{green}{}}
\newcommand{\yasu}[1]{\textcolor{orange}{}}
\newcommand{\greg}[1]{\textcolor{blue}{}}

\newcommand{\mysubsubsection}[1]{\vspace{0.1cm} \noindent {\bf #1}:}

\newcommand{\etal}{\textit{et al}.}

\begin{document}
\pagestyle{headings}
\mainmatter
\def\ECCVSubNumber{678}  

\title{Vectorizing World Buildings: Planar Graph Reconstruction by Primitive Detection and Relationship Inference} 


\titlerunning{Vectorizing World Buildings}
%
\author{First Author\inst{1}\orcidID{0000-1111-2222-3333} \and
Second Author\inst{2,3}\orcidID{1111-2222-3333-4444} \and
Third Author\inst{3}\orcidID{2222--3333-4444-5555}}

\author{Nelson Nauata\and Yasutaka Furukawa}
\authorrunning{N. Nauata et al.}
%
\institute{Simon Fraser University, BC, Canada\\
\email{\{nnauata,furukawa\}@sfu.ca}}

\maketitle

\begin{figure}
    \centering
    \includegraphics[width=\linewidth]{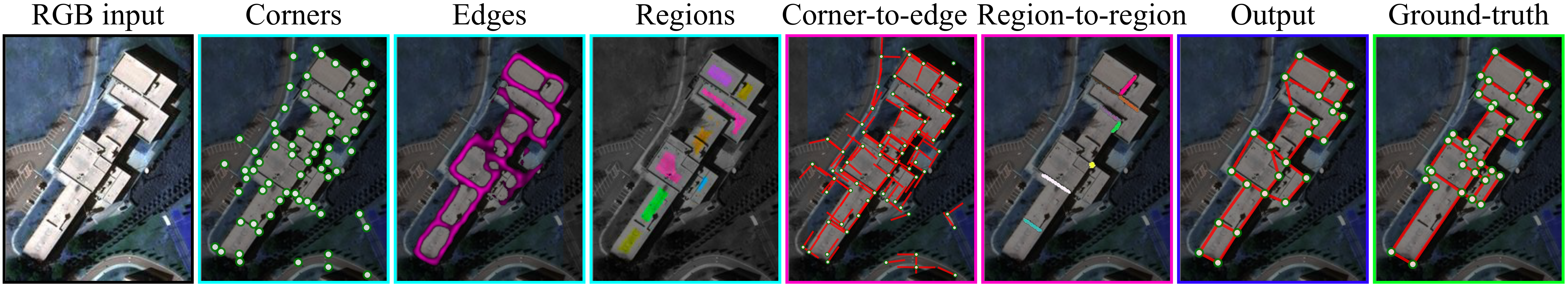}
    \caption{The paper takes a RGB image, detects three geometric primitives (i.e., corners, edges, and regions), infers their relationships (i.e., corner-to-edge and region-to-region), and fuses the information via Integer Programming to reconstruct a planar graph.}
    \label{fig:teaser}
    \vspace{-1cm}
\end{figure}

\input{sections/0_abstract.tex}
\input{sections/1_introduction.tex}

\input{sections/2_related_work.tex}

\input{sections/3_problem.tex}
\input{sections/4_method.tex}
\input{sections/5_experimental_results.tex}
\input{sections/6_conclusion.tex}
\clearpage
%
%
\bibliographystyle{splncs04}
\bibliography{egbib}
\end{document}

%% file: sections/0_abstract.tex
\begin{abstract}

This paper tackles a 2D architecture vectorization problem, whose task is to infer an outdoor building architecture as a 2D planar graph from a single RGB image. We provide a new benchmark with ground-truth annotations for 2,001 complex buildings across the cities of Atlanta, Paris, and Las Vegas. We also propose a novel algorithm utilizing 1) convolutional neural networks (CNNs) that detects geometric primitives and infers their relationships and 2) an integer programming (IP) that assembles the information into a 2D planar graph.
While being a trivial task for human vision, the inference of a graph structure with an arbitrary topology is still an open problem for computer vision.
%
Qualitative and quantitative evaluations demonstrate that our algorithm makes significant improvements over the current state-of-the-art, towards an intelligent system at the level of human perception.
%
We will share code and data.
\keywords{vectorization, remote sensing, deep learning, planar graph}
\end{abstract}

%% file: sections/1_introduction.tex
\section{Introduction}

Human vision has a stunning perceptual capability in inferring geometric structure from raster imagery. What is 
remarkable is the holistic nature of our geometry perception. 
Imagine a task of inferring a building structure as a 2D graph from a satellite image
(See Fig.~\ref{fig:teaser}).
We learn structural patterns from examples quickly, utilize the learned patterns to augment the reconstruction process from incomplete data.




\begin{figure*}[!t]
      \centering
     \includegraphics[width=\linewidth]{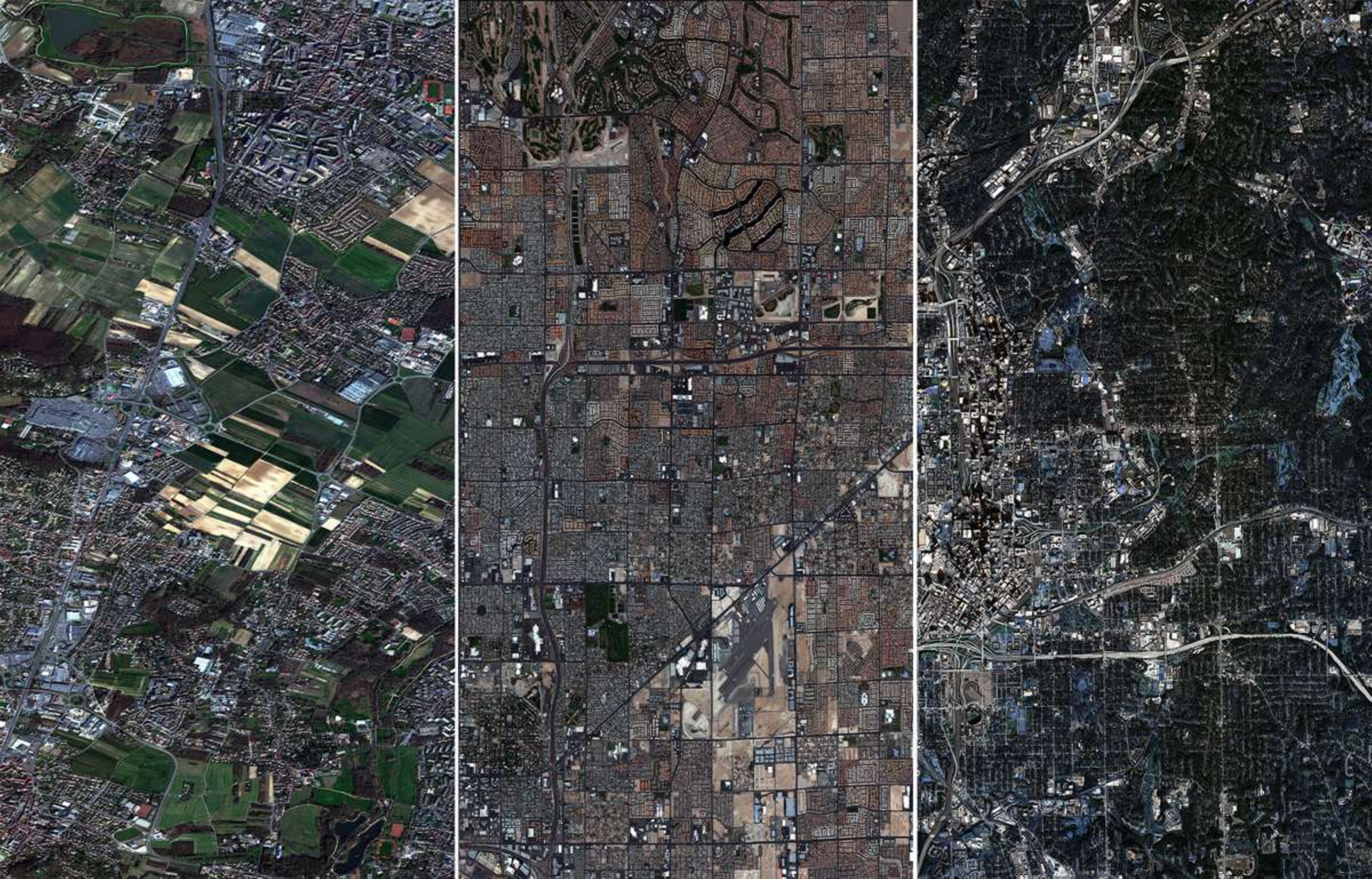}
      \caption{Sample satellite images used in our benchmark for the cities of Paris, Las Vegas and Atlanta (left-to-right). This image is part of the SpaceNet \cite{spacenet,spacenet_challenge} corpus and is hosted as an Amazon Web Services (AWS) Public Dataset.}
      \label{fig:raw_satellite}
    \end{figure*}

Computer Vision is still at its infancy in holistic reasoning of geometric structure. 
For low-level geometric primitives such as corners~\cite{liu2018floornet} or junctions~\cite{wireframe_cvpr18}, Convolutional Neural Networks (CNNs) have been an effective detector.
%
%
Unfortunately, the task of high-level geometry reasoning, for example, the construction of CAD-quality geometry, is often only possible by the hands of expert modelers. CAD-level building reconstruction at a city-scale would enable richer architectural modeling and analysis across the globe, opening doors for broad applications in digital mapping, architectural study, or urban visualization/planning.


In an effort towards more holistic structured reconstruction techniques, this paper proposes a new 2D outdoor architecture vectorization problem, whose task is to reconstruct a 2D planar graph of outdoor building architecture from a single RGB image. 
While building segmentation 
from a satellite image has been a popular problem~\cite{spacenet_challenge}, their task is to extract only the external boundary as a 1D polygonal loop. Our problem seeks to reconstruct a planar graph of an arbitrary topology, including internal building feature lines that separate roof components and yield more high-fidelity building models (See Fig.~\ref{fig:teaser}). The inference of graph topology is the challenge in our problem, which is exacerbated by the fact that
buildings on satellite images do not follow the Manhattan 
geometry due to the foreshortening effects through perspective projection.

Our approach combines CNNs and integer programming (IP) to tackle the challenge. CNNs extract low- to mid-level topology information, in particular, detecting three types of geometric primitives (i.e., corners, edges, and regions) and inferring two types of pairwise primitive relationships (i.e., corner-to-edge and region-to-region relationships). IP consolidates all the information and reconstructs a planar graph.
%

We downloaded high-resolution satellite images from SpaceNet~\cite{spacenet} corpus and annotated 2,001 complex buildings across the cities of Atlanta, Paris and Las Vegas as 2D polygonal graphs including internal and external architectural feature lines (See Fig.~\ref{fig:raw_satellite}). Our qualitative and quantitative evaluations demonstrate significant improvements in our approach over the competing methods.

In summary, the contribution of the paper is two-fold: 1) A new outdoor architecture reconstruction problem as a 2D planar graph with a benchmark;
2) A hybrid algorithm combining primitive detectors, their relationship inference, and IP, which makes significant improvements over the existing state-of-the-art.
We will share our code and data to promote further research.

%% file: sections/2_related_work.tex
\section{Related work}
Architectural reconstruction has a long history in Computer Vision.
We first review building footprint extraction methods then focus our description on vector-graphics reconstruction techniques.

\mysubsubsection{Building footprint extraction}
In the SpaceNet Building Footprint Extraction challenge~\cite{spacenet_challenge}, a ground-truth building is represented as a set of pixels, ignoring the underlying vector graphics building structure.
%
The winning method by Hamaguchi~\etal~\cite{hamaguchi2018building}
utilizes a multi-task U-Net for segmenting roads and buildings of different sizes,
producing a binary building segmentation mask.
Cheng \textit{et al.}~\cite{cheng2019darnet} utilizes CNNs for defining energy maps and optimizing polygon-based contours 
for building footprints. 
%
Acuna and Ling \etal~\cite{acuna2018efficient} formulates the footprint extraction as the boundary tracing problem, finding a sequence of vertices forming a polygonal loop. Their method is designed for general object segmentation and tends to generate too many vertices.
All these methods only extract the building footprint (i.e., external boundary) and ignores internal architectural feature lines.


\mysubsubsection{Low-level reconstruction}
Harris corner detection~\cite{szeliski2010computer}, Canny edge detection~\cite{canny1986computational}, and LSD line segment extractor~\cite{von2012lsd} are popular traditional methods for low-level geometry detection. 
More recently, deep neural network (DNN) based approaches have been an active area of research~\cite{cao2018openpose}.
%
By classifying the combination of incident edge directions, DNNs are also effective for the junction detection~\cite{wireframe_cvpr18}.
%





\mysubsubsection{Mid-level reconstruction}
Room layout estimation infers a graph of architectural feature lines from a single image, where nodes are room corners and edges are wall boundaries. Most approaches assume that the room shape is a 3D box, then solves an optimization problem with hand-engineered cost functions~\cite{hedau2009recovering,schwing2012efficient,lee2017roomnet,chao2013layout}.
%
For a room beyond a box shape, Markov Random Field (MRF) infers detailed architectural structures~\cite{furukawa2009manhattan} and Dynamic Programming (DP) searches for an optimal room shape~\cite{flint2010dynamic,flint2011manhattan}, again via hand-engineered cost functions.
%

\mysubsubsection{High-level reconstruction (knowledge)}
Given a prior knowledge about the overall geometric structure, corner detection alone suffices to reconstruct a complex graph structure. Human pose estimation is one of the most successful examples, where DNN is trained to detect human junctions with body types such as heads, right arms, and left legs~\cite{cao2017realtime}. Their connections come from a prior knowledge (e.g. right hand is connected to right arm).

\mysubsubsection{High-level reconstruction (optimization)}
A classical approach for CAD-quality 3D reconstruction is to inject domain knowledge as ad-hoc cost functions or processes in the optimization formulation~\cite{lin2013semantic}.
%
The emergence of deep learning enabled robust solutions for low-level primitive detection. However, mid to high level geometric reasoning still relies on hand-crafted optimization~\cite{liu2017raster2vec,liu2018floornet}.
%

Floor-SP~\cite{floorsp_jiacheng_iccv2019} is the closest to our work, utilizing CNN-based corner, edge, and region detection with a sophisticated optimization technique to reconstruct floorplans. However, their method suffers from two limitations. First, Floor-SP does not allow any mistake in the region detection.~\footnote{Rooms are regions in their problem and can be detected easily. Our regions are roof segments and much less distinguishable.} Second, Floor-SP requires principal directions and mostly Manhattan scenes, which is hardly true in our problem due to severe foreshortening effects. 
Our approach handles non-Manhattan scenes and utilizes region detection as soft-constraints.


\mysubsubsection{High-level reconstruction (shape-grammar)}
A shape grammar defines rules of procedural shape generation~\cite{parish2001procedural}. Procedural reconstruction exploits the shape grammar in constraining the reconstruction process.
Rectified building facade parsing is a good example,
where heuristics and hand-engineered cost functions control the process~\cite{facade-1,facade-2}.
More recently, DNNs learn to drive the procedural reconstruction for building facades~\cite{nishida2018procedural} or top-down residential houses~\cite{Zeng_2018_ECCV}.
However, these shape grammars are too restrictive and do not 
scale to more complex large buildings in our problem.

%% file: sections/3_problem.tex

\section{2D architecture vectorization problem}


    This paper proposes a new building vectorization problem, where a building is to be reconstructed as a 2D planar graph from a single RGB image.
    %
    We retrieved high-resolution satellite images from the SpaceNet~\cite{spacenet} corpus, which are hosted as an Amazon Web Services Public Dataset~\cite{spacenet_challenge} (See Fig.~\ref{fig:raw_satellite}). 
    
     \begin{figure}
      \centering
     \includegraphics[width=\linewidth]{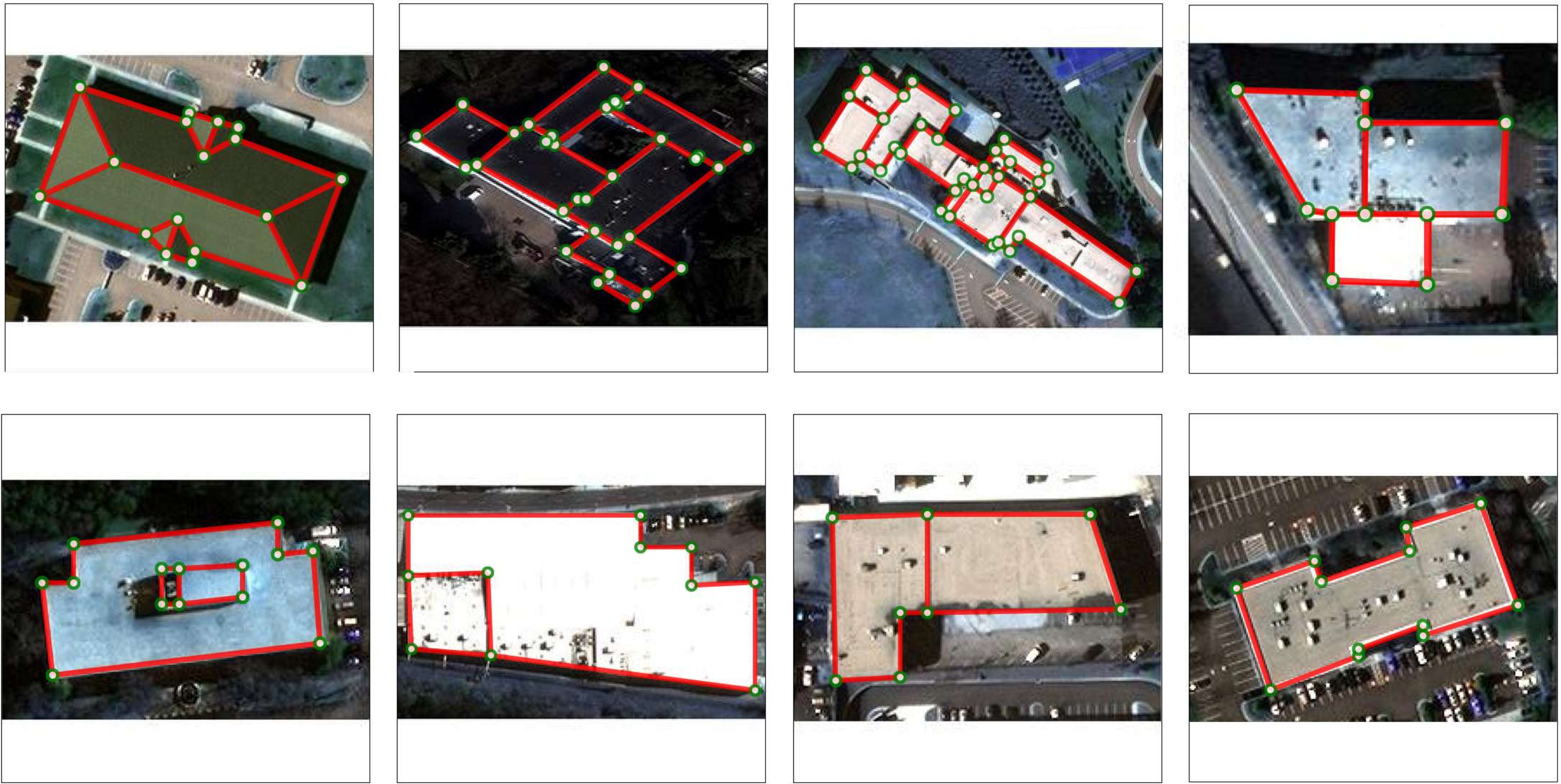}
      \caption{Sample input RGB images and their corresponding planar graph annotations.}
      \label{fig:gt_samples}
    \end{figure}
       
    We annotated 
    %
    2D planar graphs for 1010, 670, and 321 buildings from the cities of Atlanta, Paris, and Las Vegas, respectively. The average and the standard deviation of the number of corners, edges, and regions are 12.56/8.23, 14.15/9.53 and 2.8/2.19, respectively. 
    Roughly 60\% of the buildings have either 1 or 2 regions. 30\%  have 3 to 10 regions. The remaining 10\% have more than 10 regions.
    We randomly chose 1601 training and 400 testing samples.
    We refer to the supplementary document for the complete distribution of samples per number of corners, edges and regions.
    Note a region is a space bounded by the edges, which is well-defined in our planar graphs.
    When multiple satellite images cover the same city region, we chose the one with the least off-Nadir angle to minimize the foreshortening effects.
    %


    For each building instance, we crop a tight axis-aligned bounding-box with 24 pixels margin, and paste to the center of a $256\times 256$ image patch. The white color is padded at the background. We apply uniform shrinking if the bounding box is larger than $256\times 256$.
    %
    Figure~\ref{fig:gt_samples} shows sample building annotations. 
    We borrow the metrics introduced for the floorplan reconstruction~\cite{floorsp_jiacheng_iccv2019} (except for room++ metric), measuring the precision, recall, and f1-scores for the corner, edge, and region primitives.~\footnote{In short, a corner is declared to be correct if there exists a ground-truth corner within a certain distance. An edge is declared to be correct if both corners are declared to be correct. A region is declared to be correct if there exists a ground-truth region with more than 0.7 IOU.
    Our only change is to tighten the distance tolerance on the corner detection from 10 pixels to 8 pixels.}

%% file: sections/4_method.tex
\section{Algorithm}
\begin{figure*}[!tb]
  \centering
  \includegraphics[width=\linewidth]{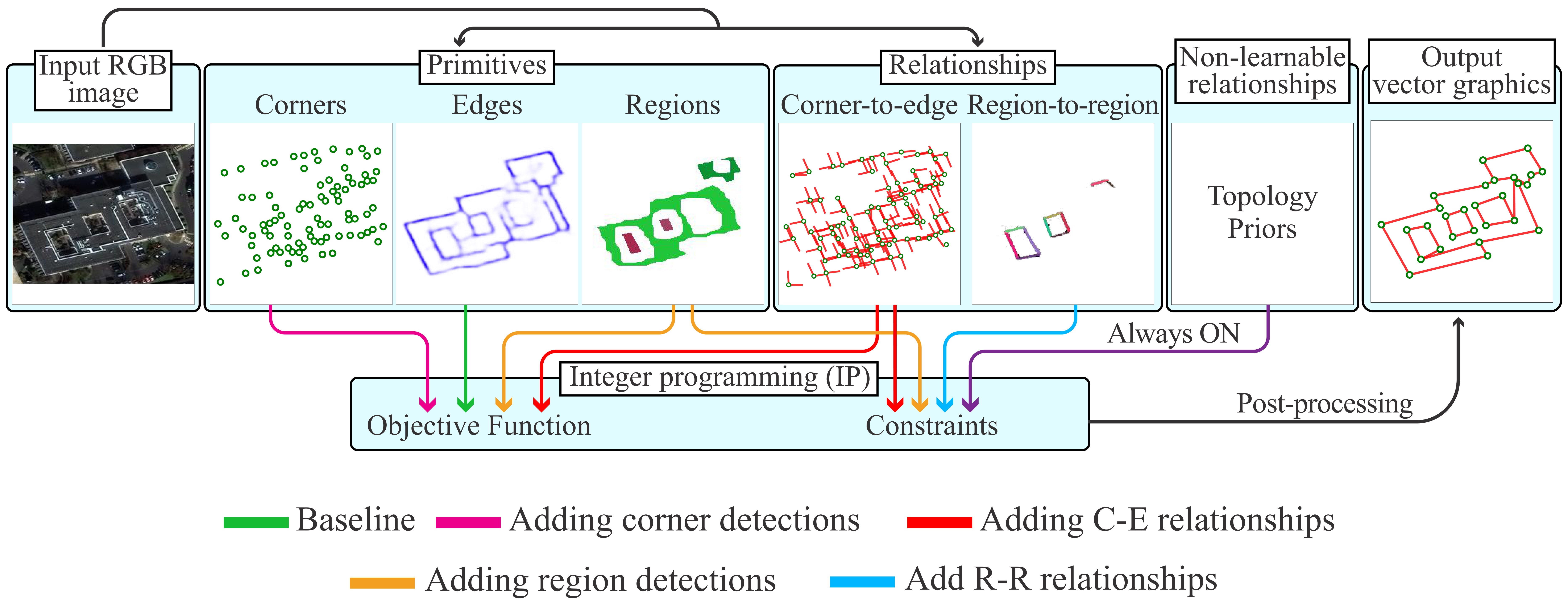}
  \caption{System overview. Our pipeline detects geometric primitives, infers their relationships, and fuses all the information into integer programming to reconstruct a planar graph.
  }
  \label{fig:overview}
\end{figure*}

Our architecture vectorization algorithm consists of three modules: CNN-based primitive detection, CNN-based primitive relationship inference, and IP optimization (See Fig.~\ref{fig:overview}). We now explain these three modules.

\subsection{Primitive detection} \label{primitive_detection}
\begin{figure}[!t]
  \centering
  \includegraphics[width=\linewidth]{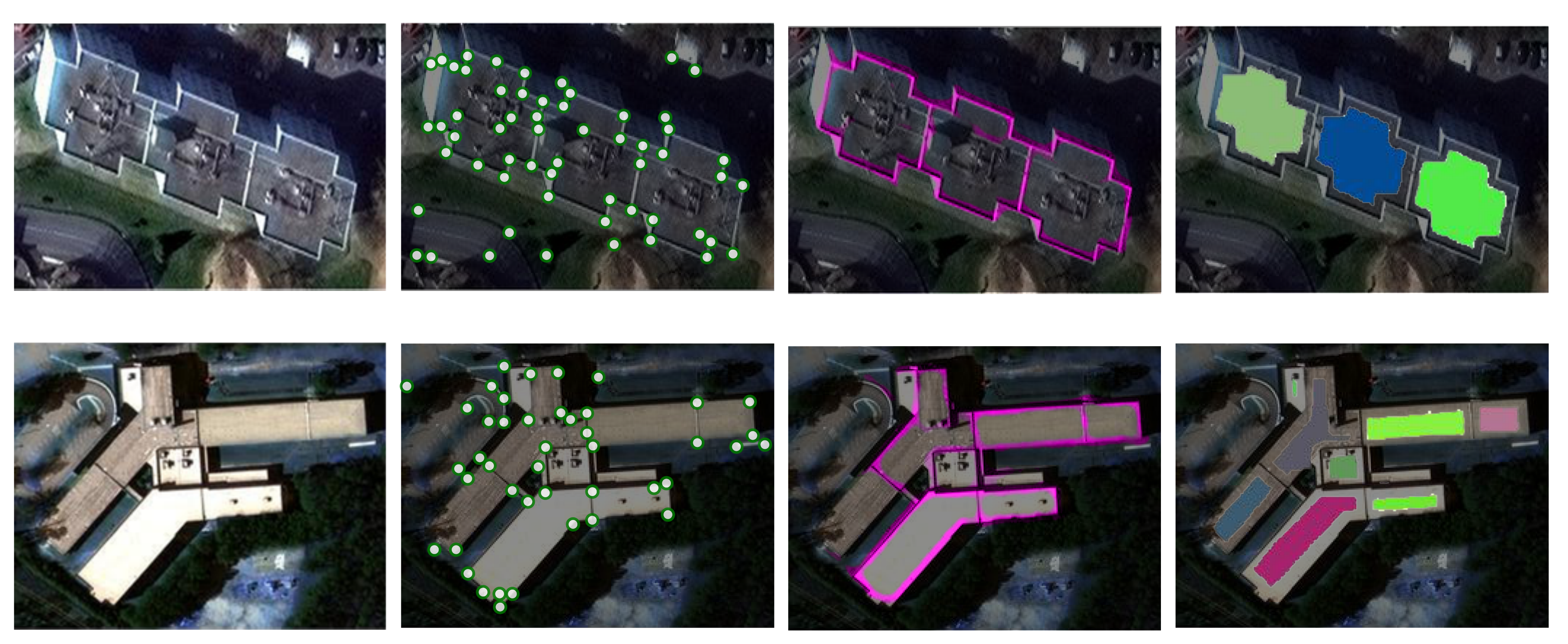}
  \caption{Primitive detection. From left to right, the figure shows an input RGB image and its corner, edge (as pixel-wise confidence map) and region detections.
  }
  \label{fig:geometrics_primitives_info}
\end{figure}

We follow Floor-SP~\cite{floorsp_jiacheng_iccv2019} and obtain corner candidates, an edge confidence image, and region candidates by standard CNN architecture (See Fig.~\ref{fig:geometrics_primitives_info}), in particular,  Fully Convolutional Network (FCN) for corners \cite{wireframe_cvpr18}, Dilated Residual Networks (DRN) \cite{Yu2017DilatedRN} for edges, and Mask-RCNN \cite{he2017mask} for regions. Corner detections are thresholded at 0.2, where the confidence scores will also be used in the optimization. Edge information is estimated as a pixel-wise confidence score without thresholding. Every pair of corner candidates is considered to be an edge candidate. Region detections are thresholded at 0.5.
We refer the full architectural specification to the supplementary document.

\begin{figure}[!t]
  \centering
  \includegraphics[width=\linewidth]{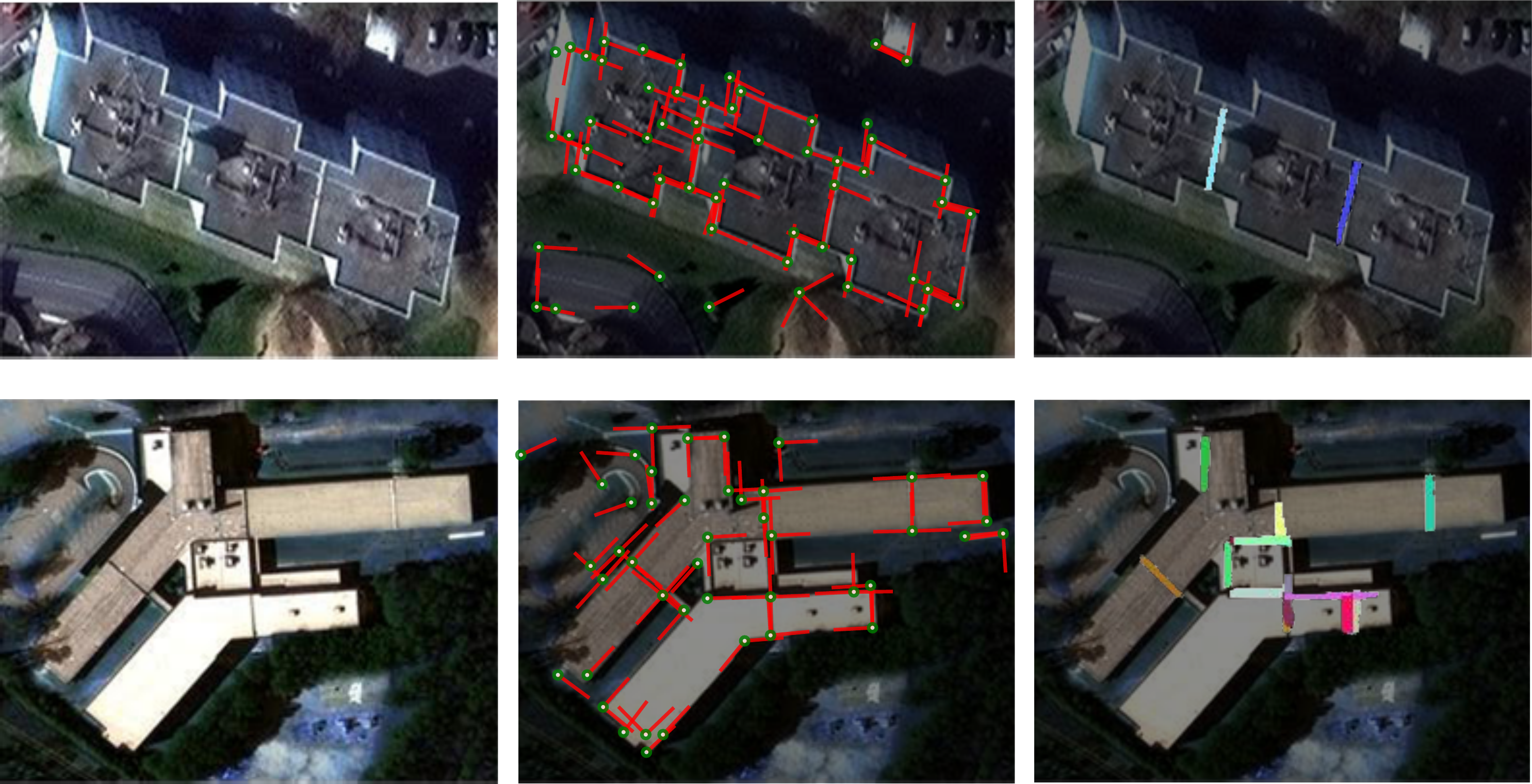}
  \caption{Primitive relationships. From left to right, the figure shows an input RGB image and its corner-to-edge relationships visualized as junctions and room-to-room relationships visualized as common boundaries.
  }
  \label{fig:relationships_info}
\end{figure}

\subsection{Primitive relationship classification} \label{relationship_classification}
We classify two types of pairwise primitive relationships by CNNs (See Fig.~\ref{fig:relationships_info}).


\mysubsubsection{Corner-to-edge relationships} For every pair of a corner and an incident edge, we compute the confidence score by utilizing the junction-type inference technique by Huang \etal without any changes~\cite{wireframe_cvpr18}. In short, we discretize 360 degrees around each detected corner into 15 angular bins, and add a module at the end of the corner detection head to estimate the presence of an edge in each bin. A corner to edge confidence score is simply set to the edge presence score in the corresponding bin. This score will be used by the objective function and the corner-to-edge relationship constraints in the optimization.


\mysubsubsection{Region-to-region relationships} 
Given a RGB image and a pair of regions, we use Mask R-CNN~\cite{he2017mask} to find their common boundary as a pixel-wise segmentation mask. More precisely, we represent the input as a 5-channel image, where the two regions are represented as binary masks. The output is a set of common edges of the two regions, each of which is represented as a segmentation instance. When 2 regions do not have a shared boundary, Mask R-CNN should not output any segments. Detected segments are thresholded at 0.5, which are often reliable and the confidence scores will not be used in the next optimization. The common boundaries will be used by the region-to-region relationship constraints in the optimization.


\subsection{Geometric primitive assembly via IP} \label{ip_formulation}
Integer Programming (IP) fuses detected primitives and their relationship information into a planar graph, where the inspiration of our formulation comes from the floorplan vectorization works by Liu et al.~\cite{liu2017raster2vec}.

\mysubsubsection{Objective function} Indicator variables are defined for each primitive: (1) $I_{cor}$ for a corner $c\in\mathcal{C}$; (2) $I_{edg}$ for an edge $e\in\mathcal{E}$; and (3) $I_{reg}$ for a region $r\in\mathcal{R}$. 
After the optimization, we collect the set of primitives whose indicator variables are 1 as a building reconstruction.
We also have an indicator variable $I_{dir}$ for a corner to an incident edge direction relationship. The variable becomes 1, if a corner has an incident edge along the direction (with binning).

The objective function consists of the three terms:
%
\begin{equation}\label{eqn:objective_function}
\begin{split}
\max_{\{I_{cor},I_{edg},I_{reg},I_{dir}\}} & \sum_{e \in \mathcal{E}}\underbrace{(e_{_{\mathit{conf}}}c^{\prime}_{_{\mathit{conf}}} c^{\prime\prime}_{_{\mathit{conf}}}- 0.5^3)I_{edg}(e)}_{\mbox{corner and edge primitives}}   \\
    + 0.1& \sum_{c \in \mathcal{C}} \sum_{\theta \in \mathcal{D}_{c}} \underbrace{(\theta_{_{\mathit{conf}}} c_{_{\mathit{conf}}} - 0.5^2) I_{dir}(\theta, c)}_{\mbox{corner-to-edge relationship}} \\
    +&\sum_{r \in \mathcal{R}} \underbrace{I_{reg}(r)}_{\mbox{region primitive}}.
\end{split}
\end{equation}
%
$c_{\mathit{conf}}$ and $e_{\mathit{conf}}$ denotes the confidence scores for the corner and the edge detections, respectively. $\theta_{\mathit{conf}}$ denotes the corner-to-edge relationship confidence.
With abuse of notation, $c^{\prime}$ and $c^{\prime\prime}$ denotes the end-points of an edge $e$.

The first objective term states that if an edge and its two end-points have high confidence scores (i.e., their product is at least $0.5^3$), there is an incentive to select that edge. The second term suggests to select a corner and its incident edge direction if their confidence scores are high.
The third objective term suggests to select as many regions as possible.
%

The maximization of the function is subject to four constraints, which are intuitive but require complex mathematical formulations. We here focus on explaining the ideas, while referring the details to the supplementary material.
Note that we describe constraints as hard constraints, but turn them into soft constraints via slack variables before solving the problem for robustness.
Lastly, after reconstructing a graph with IP, we apply a simple post-processing and eliminate a corner when it has two incident edges that are colinear with an error tolerance of 5 degrees.

\mysubsubsection{Topology prior constraints} 
There are domain-specific constraints.
First, if an edge is active, its two end-points must also be active.
Second, no dangling edges are allowed, and every corner must have at least two incident edges. Third, no two edges can intersect, which ensures the planarity of the reconstructed graphs.


\mysubsubsection{Region primitive constraints}
Suppose a region is selected. All the edges that intersect with the region should be off. Similarly, the region must be surrounded by edges. We take a point at the region boundary and cast a ray outwards the region. We collect all the edges that intersect with the ray and enforce that at least one edge must be on. We sample such points at every 2 pixels around the region boundary.


\mysubsubsection{Region-to-region relationship constraints}
This constraint is similar in spirit to the region primitive constraint but is more powerful. Suppose a pair of regions have a common boundary prediction as a segmentation mask. The constraint enforces that at least one edge is selected nearby the mask. To be precise, we fit a line segment to the boundary segment, consider an orthogonal line segment (16 pixels in length) at the center. We collect all the edge primitives that intersect with the last line segment, and enforce that one edge will be chosen.

\mysubsubsection{Corner-to-edge relationship constraints}
If the incident indicator is on, the corresponding corner must be on, and one of the edges in the corresponding directional bin must be on.
If two edges are incident to the same corner and within 5 degrees in angular distance, both edges cannot be on at the same time.
If corner-to-edge compatibility score from the relationship inference is below 0.2,
we do not allow any edges in that direction bin to be on.

%% file: sections/5_experimental_results.tex
\section{Experimental results}



We have implemented the proposed DNNs in PyTorch and the IP optimization in Python with Gurobi (a quadratic integer programming solver). We have used a workstation with Intel Xeon processors (2.2GHz) and NVidia GTX 1080 GPU with 11GB of RAM. The training usually takes 2 days for the primitive detectors and relationship classifiers. At test time, the network inference takes a fraction of a second, while IP normally takes less than 5 minutes, but can take up to 1 hour for some complex buildings.

\subsection{Comparative evaluations}\label{sec:comparative}
Table~\ref{table_all} shows our main result, comparing our approach against five competing methods: PolyRNN++~\cite{acuna2018efficient}, PPGNet~\cite{zhang2019ppgnet}, Hamaguchi~\cite{hamaguchi2018building}, L-CNN~\cite{zhou2019end} and Floor-SP~\cite{floorsp_jiacheng_iccv2019}.

\begingroup
\renewcommand{\arraystretch}{1.1}
\begin{table*}[tb]
\caption{\textbf{Quantitative evaluations.} $\mbox{P}_{C}$, $\mbox{P}_{E}$, and $\mbox{P}_{R}$ denote corner, edge, and region primitive information, respectively.
$\mbox{R}_{CE}$ and $\mbox{R}_{RR}$ denote corner-to-edge and region-to-region relationship information, respectively. The \textcolor{cyan}{cyan}, \textcolor{orange}{orange}, and \textcolor{magenta}{magenta} indicate the top 3 results.}
    \label{table_all}
\centering
\begin{tabular}{lccccccccc}
\toprule
Model & \multicolumn{3}{c}{Corner} & \multicolumn{3}{c}{Edge} & \multicolumn{3}{c}{Region}\\
\cmidrule(lr){2-4}\cmidrule(lr){5-7}\cmidrule(lr){8-10}
& Prec. & Recall & f1 & Prec. & Recall & f1 & Prec. & Recall & f1\\
\midrule
PolyRNN++ \cite{acuna2018efficient} & 49.6 & 43.7 & 46.4 & 19.5 & 15.2 & 17.1 & 39.8 & 13.7 & 20.4 \\
PPGNet \cite{zhang2019ppgnet} & 78.0 & \textcolor{orange}{69.2} & 73.3 & 55.1 & \textcolor{orange}{50.6} & 52.8 & 32.4 & 30.8 & 31.6 \\
Hamaguchi \etal \cite{hamaguchi2018building} & 58.3 & 57.8 & 58.0 & 25.4 & 22.3 & 23.8 & 51.0 & 36.7 & \textcolor{magenta}{42.7} \\
L-CNN \cite{zhou2019end} & 66.7 & \textcolor{cyan}{86.2} & \textcolor{orange}{75.2} & 51.0 & \textcolor{cyan}{71.2} &\textcolor{cyan}{59.4} & 25.9 & \textcolor{magenta}{41.5} & 31.9 \\
Floor-SP \cite{floorsp_jiacheng_iccv2019} & 55.0 & 51.4 & 53.1 & 29.0 & 26.9 & 27.9 & 39.0 & 32.5 & 35.5 \\
\hline
Ours ($\mbox{P}_{E}$) & 75.0 & 41.5 & 53.4 & 52.4 & 15.6 & 24.1 & 66.7 & 0.5 & 1.0 \\
Ours ($\mbox{P}_{E}+\mbox{P}_{C}$) & \textcolor{magenta}{85.3} & 57.9 & 69.0 & \textcolor{magenta}{66.8} & 29.8 & 41.2 & \textcolor{cyan}{81.6} & 6.9 & 12.6 \\
Ours ($\mbox{P}_{E}+\mbox{P}_{C}+\mbox{R}_{CE}$) & 81.3 & \textcolor{magenta}{66.1} & 72.9 & 62.5 & 38.8 & 47.9 & \textcolor{magenta}{71.7} & 15.6 &  25.6 \\
Ours ($\mbox{P}_{E}+\mbox{P}_{C}+\mbox{R}_{CE}+\mbox{P}_{R}$) & \textcolor{cyan}{91.7} & 61.6 & \textcolor{magenta}{73.7} & \textcolor{orange}{68.0} & 44.2 & \textcolor{magenta}{53.6} & \textcolor{orange}{71.8} & \textcolor{orange}{46.6} & \textcolor{orange}{56.5} \\
Ours ($\mbox{P}_{E}+\mbox{P}_{C}+\mbox{R}_{CE}+\mbox{P}_{R} +\mbox{R}_{RR}$) & \textcolor{orange}{91.1} & 64.6 & \textcolor{cyan}{75.6} & \textcolor{cyan}{68.1} & \textcolor{magenta}{48.0} & \textcolor{orange}{56.3} & 70.9 & \textcolor{cyan}{53.1} & \textcolor{cyan}{60.8} \\
\bottomrule
\end{tabular}
\end{table*}
\endgroup

\begin{figure}[!p]
    \centering
    \includegraphics[width=\textwidth]{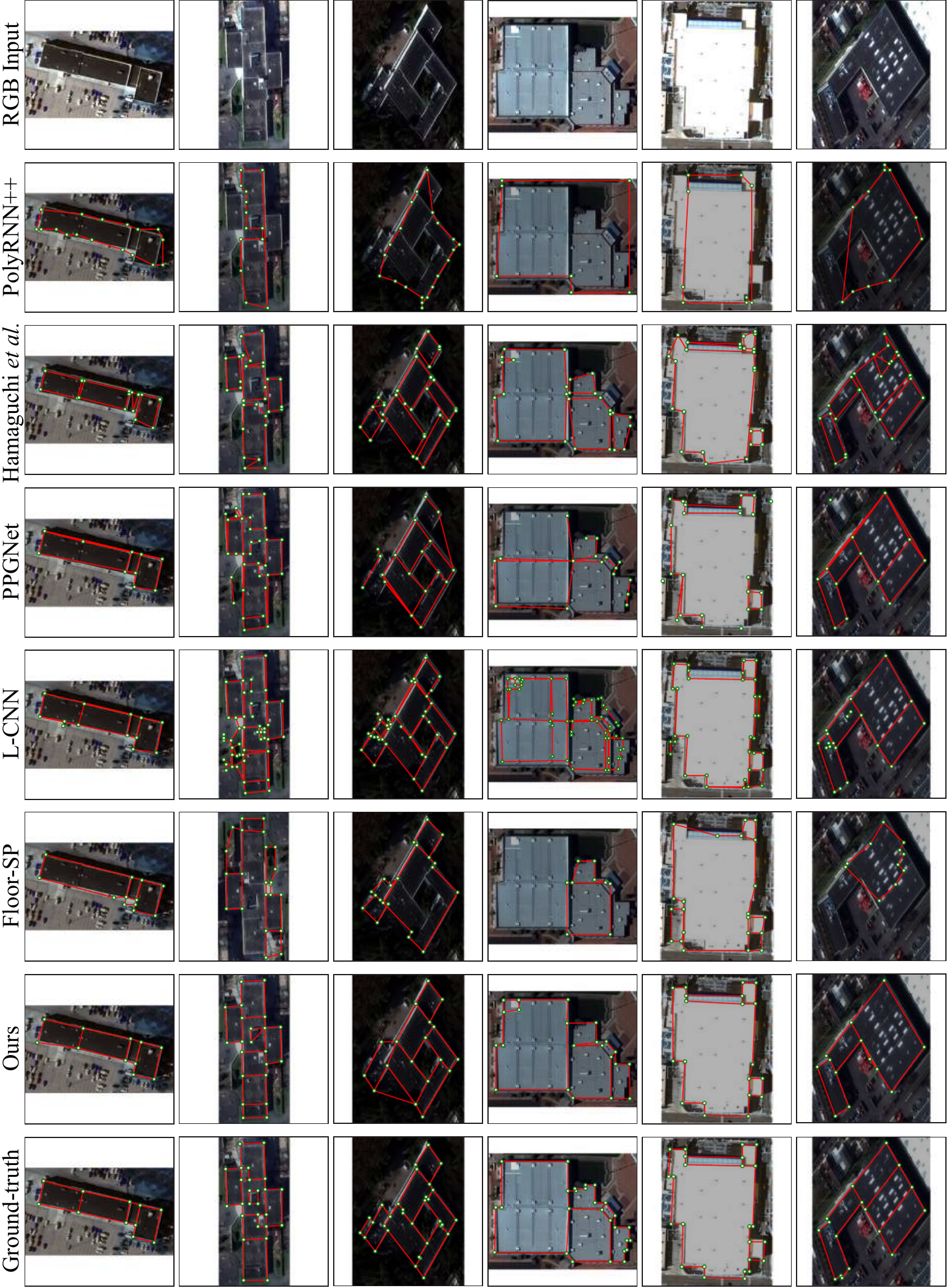}
    \caption{Comparisons against five competing methods. The top row is the input image. The last row is the ground-truth.
    The supplementary document contains results for the entire testing set.}     \label{fig:comparison_all}
\end{figure}

\begin{figure}[tb]
\centering
\includegraphics[width=\linewidth]{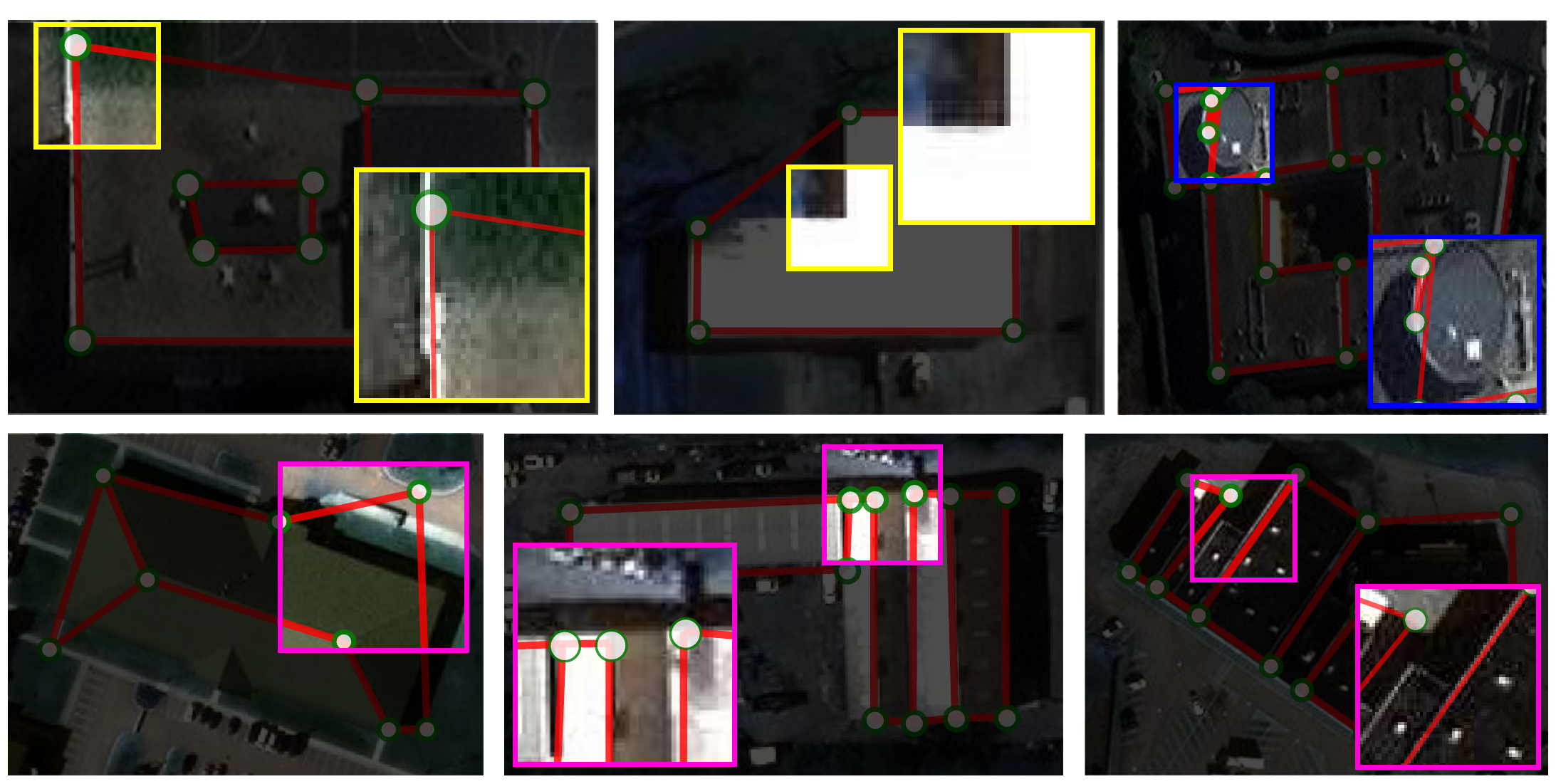}
\caption{Three major failure modes from our method: missed corner detections (yellow), curved buildings (blue), and weak image signals (magenta).} 
\label{fig:failure}
\end{figure}

\begin{figure}[tb]
    \centering
    \includegraphics[width=\textwidth]{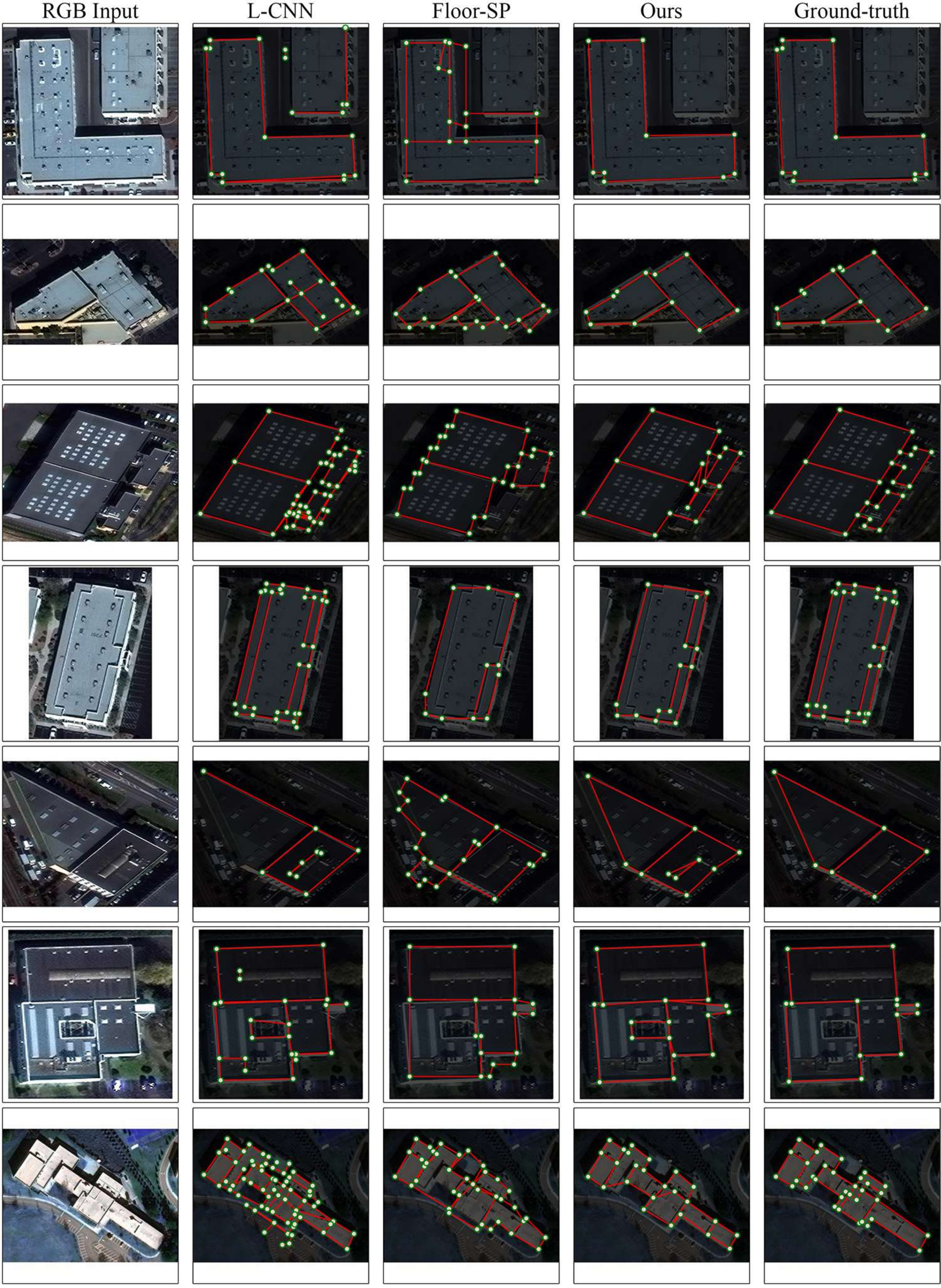}
    \caption{More detailed comparisons against L-CNN~\cite{zhou2019end} and Floor-SP~\cite{floorsp_jiacheng_iccv2019}.
    }
    \label{fig:ours_vs_floor_sp}
\end{figure}

\begin{figure}[!p]
    \centering
    \includegraphics[width=\textwidth]{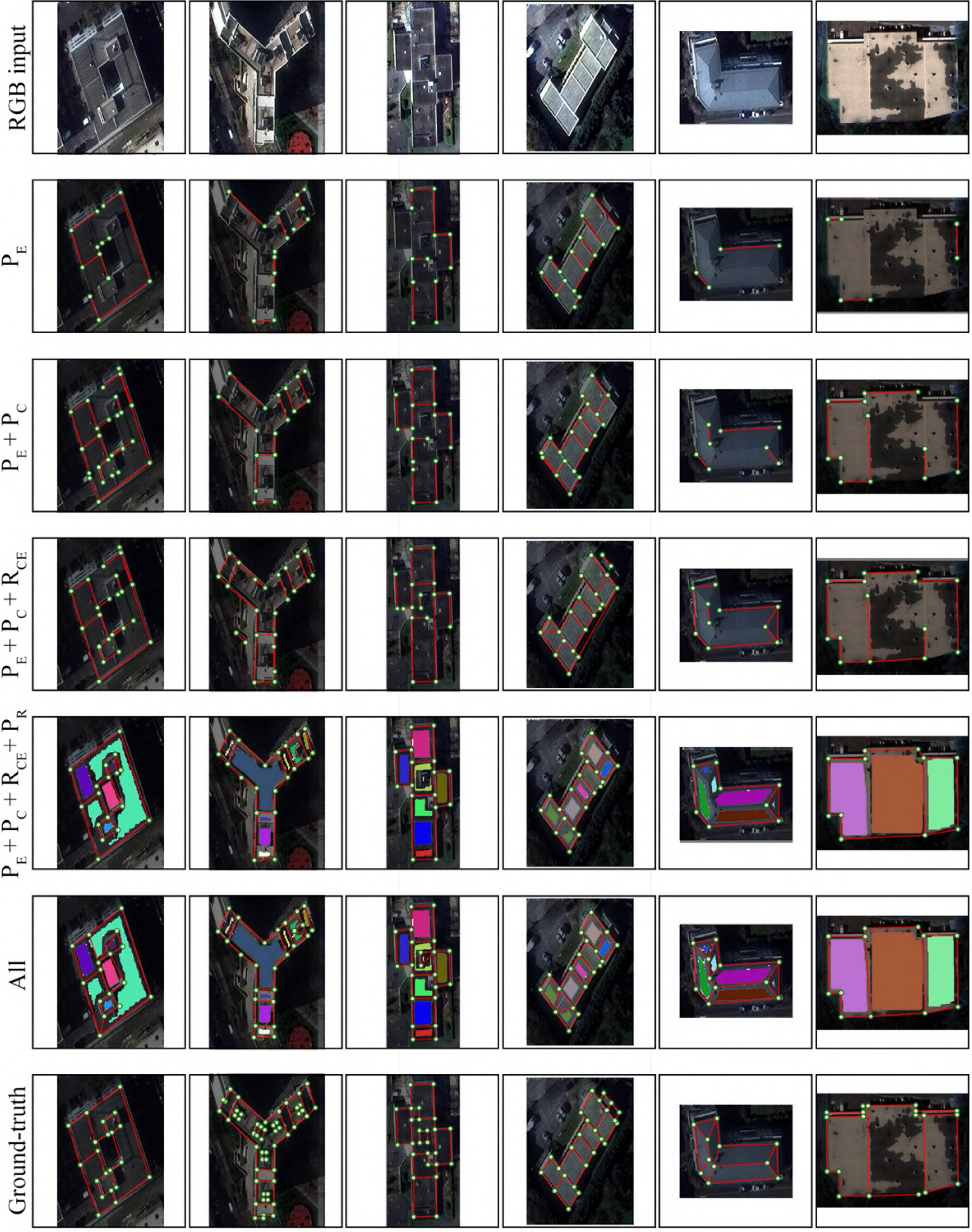}
    \caption{Qualitative evaluation for ablation study. Figure shows results for input RGB image displayed in the first row for ablation experiments presented in Section~\ref{sec:ablation} (same order) for multiple target buildings (columns) followed by ground-truth in the last row.}
    \label{fig:results}
\end{figure}

\vspace{0.1cm}

\noindent $\bullet$ PolyRNN++ traces the building external boundary in a recurrent fashion~\cite{acuna2018efficient}. We fine-tuned all their released pretrained models, in particular, ``Recurrent Decoder plus Attention'', ``Reinforcement Learning'', ``Evaluator Network'', and ``Gated Graph Neural Network''. However, we found that fine-tuning only ``Recurrent Decoder plus Attention'' achieved the best results and is used in our evaluation.

\noindent $\bullet$ PPGNet~\cite{zhang2019ppgnet} and L-CNN~\cite{zhou2019end} were reproduced by simply taking the official code and training on our data.

\noindent $\bullet$ Hamaguchi~\cite{hamaguchi2018building} won the SpaceNet Building Footprint Extraction challenge~\cite{spacenet_challenge}. 
The authors graciously trained their model and produced results using our data.
Since their method produces pixel-wise binary masks of building footprints~\cite{hamaguchi2018building}, which performs poorly in our metrics,
%
%
we utilized the OpenCV implementation of the Ramer-Douglas-Peucker algorithm with a threshold of 10 to simplify the boundary curve.

\noindent $\bullet$ Floor-SP~\cite{floorsp_jiacheng_iccv2019} is a state-of-the-art floorplan reconstruction system. Their algorithm is sensitive to the principal direction extraction (PDE), which becomes challenging against severe foreshortening effects in our problem. We tried to improve their PDE implementation without  much success and used their default code, which extracts a mixture of 4 Manhattan frames (8 directions).



%
%

\vspace{0.1cm}
All the models were trained and tested on the same split. In the table, the last row (our system with all the features) has the best f1-scores for the corner and the region metrics, and the second best f1-score for the edge metric. Overall, our model makes steady improvements over the competing methods when more features are added, especially on the region metric, which is the most challenging and consistent with the visual reconstruction quality.

PPGNet and L-CNN achieve compelling f1-scores for the corners and the edges. L-CNN even outperforms our method for the edge f1-score. However,
this metric is not a good indicator as illustrated in Figures~\ref{fig:comparison_all} and \ref{fig:ours_vs_floor_sp}. 
The figures show that the reconstruction results by PPGNet and L-CNN ``look'' reasonable at first sight. However, close examinations reveal that their results suffer from thin triangles, self-intersecting edges (i.e., the graph is not actually planar), and colinear edges. Their limitation comes from the fact that they infer edges independently. Their region metrics are far behind ours, which requires more holistic structure reasoning.

Floor-SP, on the other hand, performs poorly on all the metrics. There are two reasons. First, their shortest path algorithm at the core requires accurate principal directions, whose extraction is difficult without the Manhattan constraints in our problem. Second, they assume region detections to be 100\% correct and cannot recover from region detection mistakes. As a result, the method often generates too many extraneous corners or completely miss parts of the graphs. 




 
\subsection{Ablation study}\label{sec:ablation}
The bottom half of Table~\ref{table_all} and Fig.~\ref{fig:results} verify the contributions of various components in our system: the three primitive detections and two relationship inference. We use symbols $\mbox{P}_{C}, \mbox{P}_{E},$ and $\mbox{P}_{R}$ to denote if the corner, edge, and region primitive detections are used by our system. Similarly, $\mbox{R}_{CE}$ and $\mbox{R}_{RR}$ denote if the corner-to-edge and region-to-region relationships are used by our system.

\mysubsubsection{Edge detections only ($\mbox{P}_{E}$)}
Our first baseline utilizes only the edge detection results, that is, seeking to maximize ($\sum_{e \in \mathcal{E}} (c^{e} - 0.5) I_{edg}(e)$). In short, this baseline accepts all the edges whose score are above 0.5.

\mysubsubsection{Adding corner detections ($\mbox{P}_{E}, \mbox{P}_{C}$)}
The second baseline ($\mbox{P}_{E}+\mbox{P}_{C}$) adds the corner detection results, seeking to maximize 
$\sum_{e \in \mathcal{E}} (e_{_{conf}} c^{\prime}_{_{conf}} c^{\prime\prime}_{_{conf}} - 0.5^3) I_{edg}(e)$. 
This baseline accepts all the edges, whose scores based on the corner and the edge detection are above $0.5^3$. This effectively suppresses the corner and edge false positives, noticeably improving the precision and recall for both primitives.

\mysubsubsection{Adding C-E relationships
($\mbox{P}_{E},\mbox{P}_{C},\mbox{R}_{CE}$)}
This baseline adds the corner-to-edge relationship constraints and the corresponding objective in Eq. \ref{eqn:objective_function} to the formulation, enforcing the corner and edge variables to follow the predicted relationships. 
This change alone doubles the region f1-score.

\mysubsubsection{Adding region detections ($\mbox{P}_{E},\mbox{P}_{C},\mbox{R}_{CE},\mbox{P}_{R}$)}
With the addition of region detections, this baseline has the complete objective function, while the region constraints are also added. This baseline  allows IP to conduct high-level geometry reasoning, and brings significant boost to the region metrics.

\mysubsubsection{Adding R-R relationships ($\mbox{P}_{E},\mbox{P}_{C},\mbox{R}_{CE},\mbox{P}_{R},\mbox{R}_{RR}$)}
Finally, our system with all the features achieve the best results, successfully reconstructing complex large buildings.

\subsection{Failure cases}

There are three major failure modes as illustrated in Fig.~\ref{fig:failure}. First, our algorithm cannot recover from corners missed by the corner detector. Missing corners lead to missing incident graph structure or corrupted geometry.
Second, our algorithm assumes piece-wise linear structure and cannot handle curved buildings, while the system tries to approximate the shape as shown in the figure.
Third, our system also fails when the image signal becomes weak and the detected primitive and/or relationship information also become weak.

%% file: sections/6_conclusion.tex
\section{Conclusion} 

This paper introduces a novel outdoor architecture vectorization problem with a benchmark, whose task is to reconstruct a building architecture as a 2D planar graph from a single image. The paper also presents an algorithm that uses CNNs to detect geometric primitives and infer their relationships, where IP fuses all the information into a planar graph through holistic geometric reasoning.
The proposed method makes significant improvements over the existing state-of-the-art.
The growing volume of remote sensing data collected by space and airborne assets facilitates myriad of scientific, engineering, and commercial applications in geographic information systems (GIS).
%
We believe that this paper makes an important step towards the construction of an intelligent GIS system at the level of human perception.
We will share our code and data to promote further research.

\mysubsubsection{Acknowledgement}
This research is partially supported by NSERC Discovery Grants, NSERC Discovery Grants Accelerator Supplements, and DND/NSERC Discovery Grant Supplement. This research is also supported by the Intelligence Advanced Research Projects Activity (IARPA) via Department of Interior / Interior Business Center (DOI/IBC) contract number D17PC00288. The U.S. Government is authorized to reproduce and distribute reprints for Governmental purposes notwithstanding any copyright annotation thereon. The views and conclusions contained herein are those of the authors and should not be interpreted as necessarily representing the official policies or endorsements, either expressed or implied, of IARPA, DOI/IBC, or the U.S. Government.